\documentclass{article}

\usepackage{microtype}
\usepackage{graphicx}
\usepackage{subfigure}
\usepackage{booktabs} 

\usepackage{hyperref}

\usepackage[accepted]{icml2025}

\usepackage{amsmath}
\usepackage{amssymb}
\usepackage{mathtools}
\usepackage{amsthm}

\usepackage[capitalize,noabbrev]{cleveref}

\usepackage[T1]{fontenc}
\usepackage{graphicx}
\usepackage{balance}
\usepackage{latexsym}
\graphicspath{{./images/}}
\usepackage{booktabs} 
\usepackage{color}  
\usepackage{amsmath}  
\usepackage{subcaption}
\usepackage{caption}
\usepackage{tikz}
\usepackage{colortbl} 
\usepackage{framed}
\usepackage{multirow}
\usepackage{multicol}
\usepackage{url}
\usepackage{verbatim}
\usepackage{cancel}
\usepackage{xspace} 
\usepackage[moderate, leading=normal]{savetrees} 
\usepackage{bbold} 
\usepackage{arydshln}
\usepackage{float}
\usepackage{fancyvrb}  

\newcommand{\struct}[1]{\texttt{\small #1}}

\newcommand{\phrase}[1]{\textit{``#1''}}

\newcommand{\myparagraph}[1]{\noindent \textbf{#1}.}

\newcommand{\squishlist}{
	\begin{list}{$\bullet$}
		{ \setlength{\itemsep}{0pt}
			\setlength{\parsep}{3pt}
			\setlength{\topsep}{3pt}
			\setlength{\partopsep}{0pt}
			\setlength{\leftmargin}{1.5em}
			\setlength{\labelwidth}{1em}
			\setlength{\labelsep}{0.5em} } }
	\newcommand{\squishend}{
\end{list}  }

\newcommand{\cedar}{\textsc{Cedar}\xspace}

\usepackage[textsize=tiny]{todonotes}

\icmltitlerunning{CEDAR: Context Engineering for Agentic Data Science}

\begin{document}

\twocolumn[
\icmltitle{CEDAR: Context Engineering for Agentic Data Science}



\icmlsetsymbol{equal}{*}

\begin{icmlauthorlist}
\icmlauthor{Rishiraj Saha Roy}{comp}
\icmlauthor{Chris Hinze}{comp}
\icmlauthor{Luzian Hahn}{comp}
\icmlauthor{Fabian K\"uch}{comp}
\end{icmlauthorlist}

\icmlaffiliation{comp}{Department of Generative AI,
Fraunhofer IIS,
Am Wolfsmantel 33, 91058 Erlangen, Bavaria, Germany}

\icmlcorrespondingauthor{Rishiraj Saha Roy}{rishiraj.saha.roy@iis.fraunhofer.de}

\icmlkeywords{Data science, Agentic systems, Context engineering, Tool use}

\vskip 0.3in
]





\begingroup
\renewcommand\thefootnote{\arabic{footnote}}
\setcounter{footnote}{0}
\footnotetext[1]{Department of Generative AI, Fraunhofer IIS, Am Wolfsmantel 33, 91058 Erlangen, Bavaria, Germany. Correspondence: \texttt{rishiraj.saha.roy@iis.fraunhofer.de}}
\endgroup

\begin{abstract}
We demonstrate \cedar
(\underline{C}ontext \underline{E}ngineering for \underline{D}ata science with \underline{A}gent \underline{R}outing),
an application for automating data science (DS) tasks with
an agentic setup.
Solving DS problems with LLMs is an
underexplored area that has
immense market value.
The challenges are manifold:
task complexities, data sizes, computational limitations, and context restrictions.
We show that these can be alleviated via effective context engineering.
We first impose structure into the initial prompt with DS-specific input fields,
that serve as instructions
for the agentic system.
The solution is then materialized as an enumerated sequence of
interleaved plan and code blocks generated by separate LLM agents,
providing a readable structure to the context
at any step of the workflow.
Function calls for generating these intermediate texts,
and for corresponding Python code,
ensure that data stays local, and only
aggregate statistics and associated instructions
are injected into LLM prompts.
Fault tolerance and context management are introduced via iterative code generation
and smart history rendering.
The viability of our
agentic data scientist
is demonstrated using
canonical Kaggle challenges.
\end{abstract}

\section{Introduction}
\label{sec:intro}

\myparagraph{Motivation}
In traditional data science (DS), a human expert, the data scientist, writes scripts for
entire pipelines, including standard steps like 
data preprocessing, feature engineering,
hyperparameter tuning,
and finally computing metrics and visualizing insights.
However, this work is tedious and repetitive, and can be significantly optimized with modern LLMs.
For example, many users are exploring options like ChatGPT Advanced Data Analysis,
where one simply uploads data and articulates entire problems as prompts.

However, such approaches have several limitations:
(i) concrete instructions for real DS projects are more complex than can be solved via one-shot prompting;
(ii) capabilities of generative models are still limited with respect to
arbitrary mathematical computations;
(iii) data files are often very large and cannot simply be uploaded as attachments
(ChatGPT Advanced Data Analysis has a limit of 512 MB per file, many Kaggle files are much larger);
(iv) there are often privacy concerns and users may feel insecure uploading
their enterprise data
to cloud-based LLMs;
and (v) DS solutions are typically multi-step: as a solution workflow progresses, 
simply packing all instructions, text, code, data, 
and results into the running context
makes it unintelligible for most LLMs, and often exceeds context length limits.

\myparagraph{Limitations of state-of-the-art} Agentic systems
now power many real-world information retrieval (IR) and machine learning (ML) applications, 
where LLMs perform various roles in complex task pipelines
(see, for example,~\cite{he2024frontiers}, \cite{chu2025llm}, \cite{qiu2024llm}).
DS is a unique instance of such complexity,
tightly coupling IR, ML, and traditional statistics.
Agentic systems with effective context management can alleviate most of the above problems
for DS tasks.
This has become an active research area~\cite{maojun2025survey,jing2024dsbench},
where the best systems usually achieve impressive results via sophisticated prompt chaining.
But we found that it is very difficult for a typical end-user to get a clear idea 
into how the task is actually being solved~\cite{grosnit2024large,guo2024ds,hong2024data}.
The closest work to ours' is the very recent Jupyter Agent 2~\cite{jupyter} from Hugging Face.
However, contrary to our application, Jupyter Agent 2 does not run locally:
data needs to be uploaded to its cloud, and the time
taken depends heavily on one's network bandwidth (usually very slow).
Also, the privacy of the uploaded data is not guaranteed.
The purpose of our system \cedar is to 
bring transparency and simplicity into DS solutions with LLMs.

\begin{figure*} [t]
    \centering
    \includegraphics[width=\textwidth]{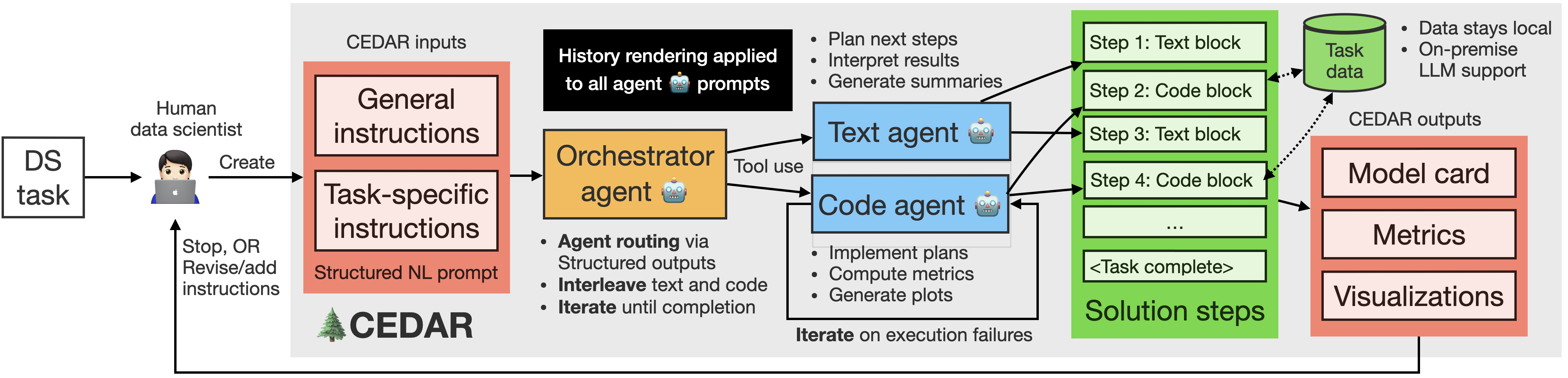}
    \caption{The CEDAR system with three LLM agents, facilitating a human data scientist's job.}
    \label{fig:overview}
\end{figure*}

\myparagraph{Approach} CEDAR relies on effective \textit{context engineering}, collectively referring to strategies for 
maintaining an optimal set of tokens during LLM inference, including all
information that may land there outside of prompts\footnote{\url{https://www.anthropic.com/engineering/effective-context-engineering-for-ai-agents}}.
An overview of our workflow is in Fig.~\ref{fig:overview}.
The data scientist formulates the task as a structured prompt, that is passed on to
an orchestrator agent. This agent routes text and code generation requests to sub-agents
as tools, to generate a readable solution with short steps. The human inspects final outputs
and revises instructions for further iterations, if necessary. 
Code and other artifacts are publicly available at \url{https://github.com/Fraunhofer-IIS/cedar}.

\myparagraph{Audience} The app can be used by beginner to expert-level IR, ML, NLP, and AI practitioners.
IR and ML basics help, but there is no prerequisite of core DS knowledge to interpret the workflow: each solution step has natural language (NL) explanations.
Beginners can get a feel of how basic data science tasks are solved.
Intermediate users can contrast the faithfulness of the generated solution to the original intent, and gain insights into how agentic systems can be built for simplifying DS.
Experts can scrutinize LLM-generated code snippets,
and whether the generated solution mimics code written by a human data scientist.

\myparagraph{Scope} Our system can solve beginner-level data science tasks from Kaggle.
The current requirements are simply to have a clearly articulated goal and well-defined data.

\section{Method}
\label{sec:method}

We now describe the key features of the \cedar pipeline:
prompts, the agentic setup, tool use, reliance on code for plan execution, and history rendering.

\subsection{Structured prompts}
\label{subsec:prompts}

To relieve the user from creating verbose prompts 
containing every pertinent task detail, we create a structured form that covers key 
aspects of most DS tasks. We split the project summary into high-level \textit{general instructions} (estimated number of solution steps, expected number of plots, verbosity of plans, etc.) and \textit{task-specific instructions} (with items like \struct{Task description}, \struct{Data description}, \struct{Data location}, \struct{Metrics}, \struct{Inputs}, \struct{Outputs}, and \struct{Special instructions}). This makes DS tasks more understandable to LLMs.

\subsection{Interleaved text and code}
\label{subsec:textcode}

Instead of expecting the LLM to directly emit a final solution,
say, like an optimized set of metrics, we make the system  
generate a human-readable workflow with succinct, numbered
steps starting from data loading and all the way to model card generation.
Each \phrase{Step} consists of a plan and a corresponding code snippet that implements this plan~\cite{wang2024executable},
like a Jupyter notebook.
We then have an output with scrutable and reusable components like plans, code snippets, data snapshots, error traces, and intermediate plots.

\subsection{Agentic setup}
\label{subsec:agentic}

Such a workflow is enabled by the use of agents, i.e. different LLM instantiations with distinct prompts. We have three agents: a main orchestrator, and sub-agents for text and code generation. The orchestrator decides whether to invoke the text/code generator (routing), or, seeing the blocks generated so far, that the \textit{task
is complete}. The calls to the sub-agents usually alternate between text and code like most
Jupyter notebooks, but special situations
may need consecutive text or code blocks. All agents receive the same context: the project summary, text and code snippets so far, and outputs of code snippet execution (see \textit{history rendering} below for more details).
They only vary in their responses: the main agent emits a sub-agent call, while
the sub-agents emit text and code as per the current state of the workflow.

\subsection{Tool calls}
\label{subsec:tools}

To prevent the orchestrator from generating free-form output that
needs to be parsed, we implement the code and text generators as distinct \textit{functions} so 
that they become \textit{tools} to be called by
the main agent\footnote{\url{https://platform.openai.com/docs/guides/function-calling}}.
This has been a special feature in
modern LLMs that highly facilitates building apps that chain LLM agents together.
Further, to prevent hallucinated outputs like \texttt{write\_code} when the correct function name
is \texttt{request\_code}, we use \textit{structured outputs} by forcing the orchestrator to output JSON as per our schema\footnote{\url{https://platform.openai.com/docs/guides/structured-outputs}}.

The communication with our tools is schema-driven, to prevent failures due to hallucinated responses by the main agent.
The difference between \phrase{spec} (for \texttt{request\_text)} and \phrase{purpose} (for \texttt{request\_code}) is intentional, and it reflects the different goals of the two downstream agents (Text vs. Code).
The Text Agent's job is to produce Markdown explanations -- human-readable commentary, analysis, or narration.
The word \phrase{spec} (short for specification) captures what the Orchestrator wants the text to talk about -- i.e., the topic or scope of the Markdown. For example:
\vspace{-1em}
\begin{Verbatim}[fontsize=\scriptsize]
{
    "action": "request_text",
    "spec": "Explain how the model will be evaluated
            and metrics that have to be computed."
}    
\end{Verbatim}
\vspace{-1em}
The Text Agent writes a clear paragraph or bullet list about evaluation metrics -- not code (we can think of \phrase{spec} as a creative or descriptive brief).
The Code Agent must instead produce executable Python code. The word \phrase{purpose} indicates the goal or intent of that code cell — i.e., what this code is meant to achieve. For example:
\vspace{-1em}
\begin{Verbatim}[fontsize=\scriptsize]
{
    "action": "request_code",
    "purpose": "Load the training and test datasets
               into pandas DataFrames." 
}    
\end{Verbatim}
\vspace{-1em}
The Code Agent now generates the corresponding Python code to achieve this (\phrase{purpose} is a functional or operational instruction).
The difference helps both downstream agents interpret the Orchestrator’s intent unambiguously:
\squishlist
    \item the Text Agent focuses on narrative goals (explain, summarize, discuss); and
    \item the Code Agent focuses on technical goals (load, process, train, evaluate).
\squishend
We could rename \phrase{spec} and \phrase{purpose} to a common field like \phrase{instruction}, but then both downstream roles would need logic to interpret that differently -- which would make the orchestration less explicit and a bit more error-prone.
The \phrase{finish} action is used when the Orchestrator decides that the notebook or analysis has reached a natural stopping point -- the model has achieved its main goal and should wrap up. The accompanying \texttt{summary\_hint} (optional) gives a short textual summary of what was achieved, key results, or next steps to note in the final cell. For example:
\vspace{-1em}
\begin{Verbatim}[fontsize=\scriptsize]
{
    "action": "finish",
    "purpose": "Baseline logistic regression trained and
               evaluated. Accuracy is about 0.72. No further
               steps required." 
}    
\end{Verbatim}
\vspace{-1em}
The finish response does not invoke any downstream LLM, but
triggers the CEDAR app logic to create a final \phrase{Finished} cell in the notebook. Use of structured outputs reduces the amount of retries to solve specific sub-problems by separating routing from tool parametrization~\cite{beurer2024guiding}. A summary of orchestrator responses is in Table~\ref{tab:sop}.

\begin{table*} 
    \centering
    \resizebox{\textwidth}{!}{
    \begin{tabular}{l l p{5cm} p{5cm} p{5cm}}
        \toprule
        \textbf{Orchestrator output}     & \textbf{Handled by}  & \textbf{Meaning}                                                                                          & \textbf{Key argument} & \textbf{Output}  \\ \toprule
        \texttt{request\_text}     & Text Agent           & Generate the next Markdown cell explaining results or outlining the plan for the next step & \texttt{spec} $\mapsto$ Describes what to talk about (topic, focus, or goal of the text)               & Explanatory Markdown block (added as a \phrase{text} cell)        \\
        \texttt{request\_code}  & Code Agent           & Generate the next Python code cell that performs a concrete task (e.g. data load, training, evaluation) & \texttt{purpose} $\mapsto$ Explains what the code should achieve                & Executable Python code block (added as a \phrase{code} cell, executed immediately)         \\ 
        \texttt{finish}                  & CEDAR App Logic      & Signal that the notebook is complete -- no more steps are needed         & \texttt{summary\_hint} $\mapsto$ Optional final note summarizing results                & Marks the session as finished, and appends a final \phrase{finish} cell          \\ \bottomrule
    \end{tabular}}
    \caption{Explaining the use of Structured Outputs for possible orchestrator responses.}
    \label{tab:sop}
\end{table*}

\subsection{Code for math and data handling}
\label{subsec:code}

Generating Python \textit{code for computations} frees us from dependence on an LLMs' math capabilities.
Another significant advantage is that
\textit{data stays local}: the code operates on local data, and only snapshots and aggregate statistics are passed on to LLMs.
We support on-premise LLMs, so that even such data digests need not leave the user's system
(in case working with sensitive data).
We allow for \textit{iterative code execution}:
when the execution of a particular code block fails, the coder tool is prompted again with
the current \textit{error trace} and a request to accordingly \textit{rewrite} the code. This step 
adds robustness and
enables
recovery from name/type mismatches and missed imports, for instance.

\myparagraph{Safety} Docker is used to execute code inside containerized runtime environments that isolate application processes from other host processes, thereby reducing potential security risks. Network access can be explicitly restricted through Docker’s networking configuration to prevent unintended data sharing or the execution of malicious external code. The execution environment is based on pre-built Docker images that include commonly used data science libraries and is extended as required.

\subsection{History rendering}
\label{subsec:history}

This module takes the
task history,
i.e., the list of all blocks so far (instructions, text, code, outputs, errors),
as input,
and converts them into a compact, LLM-friendly text summary as follows:
(i) it appends user instructions;
(ii) it numbers each text/code block (\texttt{Text \#3} or \texttt{Code \#5}) for clarity 
and appends them in full
as these are not very long
(only adding outputs and not raw code loses vital rationale that led to these outputs).
Actual \textit{context bottlenecks} are code outputs and error messages, as handled next;
(iii) for past code blocks, it adds only those that ran successfully;
(iv) for these \phrase{success} blocks, it adds only \textit{heads} of outputs as
they carry the most vital information;
(v) if the latest code block resulted in error, it adds the \textit{tail} of the traceback
instead, as it pinpoints the error most specifically;
(vi) if the history so rendered exceeds $10k$ characters (configurable), only the most recent 
$10k$ are kept.
This history is then passed on as context to the agent being invoked at any given moment.

\section{Demonstration walkthrough}
\label{sec:demo}

\begin{figure*} [t]
    \centering
    \includegraphics[width=\textwidth]{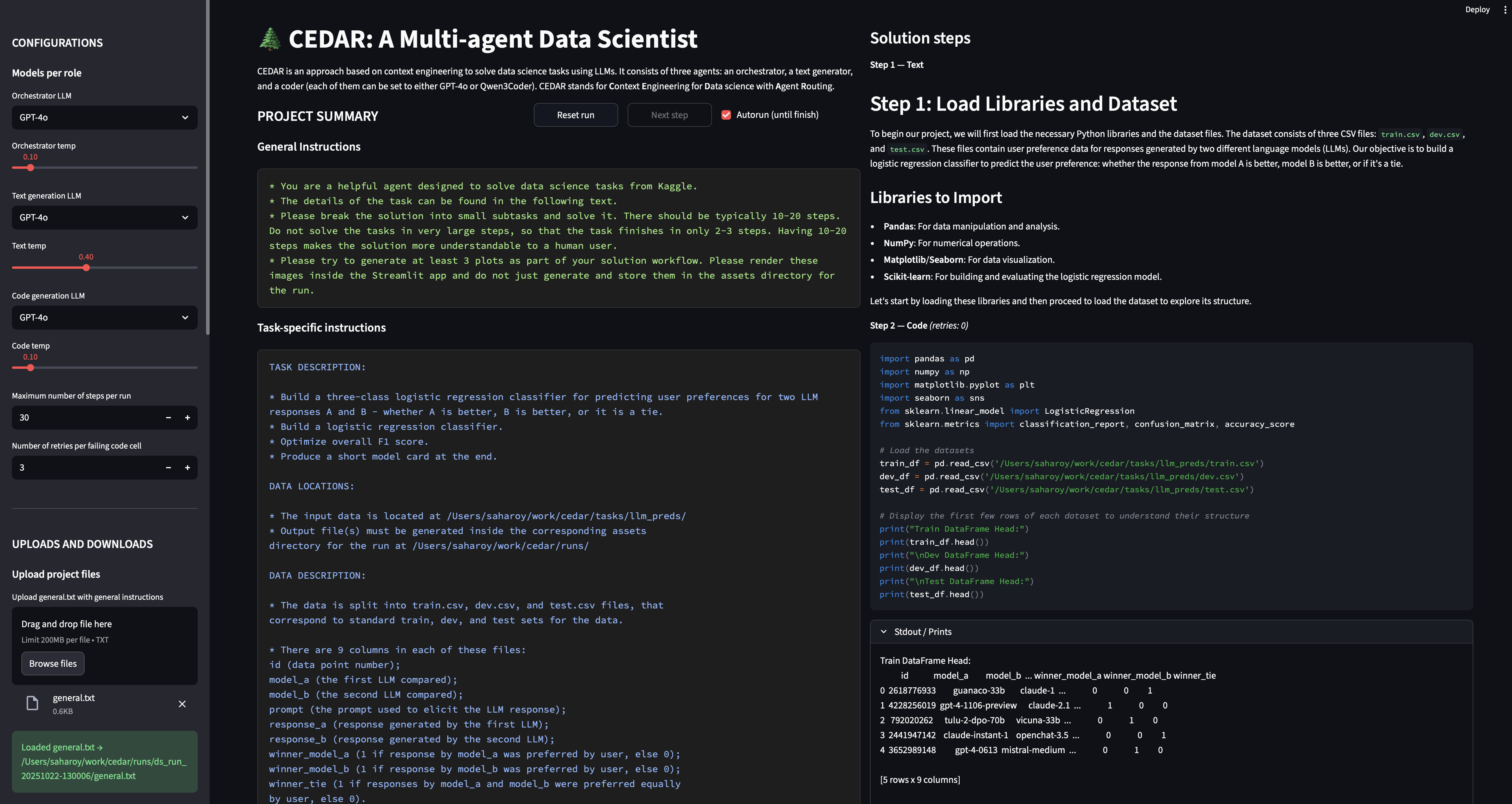}
    \caption{A screenshot of our \cedar application, solving a Kaggle competition on predicting better LLM responses.}
    \label{fig:screenshot}
\end{figure*}

\subsection{Using the app}
\label{subsec:app}

Fig.~\ref{fig:screenshot} shows \cedar
solving a canonical Kaggle competition 
on LLM fine-tuning\footnote{\url{https://kaggle.com/competitions/llm-classification-finetuning}}.
A user first selects
LLMs for orchestrator, text generator, and coder agents
(\texttt{GPT-4o}
via API,
or
\texttt{Qwen3-Coder} 30B\footnote{\url{https://arxiv.org/abs/2505.09388}}
locally via \texttt{ollama}\footnote{\url{https://ollama.com/library/qwen3-coder}}).
Qwen3-Coder was chosen as an alternative as 
it was fine-tuned specifically for agentic coding.
Next, general (green) and task-specific instructions (blue) need to be uploaded,
instantly rendered as the ``project summary''
in the broad middle panel.
The broad right panel is reserved for rendering the complete generated solution.

The user selects
between
\phrase{Next step} for inspecting a step at a time,
and \phrase{Autorun} to see the full solution
at once
($\simeq 3$ minutes for $10-20$ steps).
Steps can be \phrase{Reset} at any time.

\subsection{Assets directory}
\label{subsec:assets}

An assets' directory can be explored at any time during or
after the run: it stores the structured prompt, generated plots (also rendered online
in the app), metrics, the model card and debug logs. We allow for \textit{exporting} the solution 
as JSON, Markdown, or as a Jupyter notebook
(making the solution editable),
and \textit{importing} a saved run as JSON. 
The run can be continued from where we left off when the saved JSON for a partially completed
run is uploaded.

\subsection{Configurations and debugging}
\label{subsec:config}

The leftmost panel provides several knobs: (i) temperatures for each of
our agents (to transition between response repeatability and diversity);
(ii) maximum solution steps (default $30$,
suitable for most Kaggle tasks);
and (iii) maximum code retries
(default $3$, solves commonly observed errors).
An \textit{autorun trace} shows running summaries 
for history rendering
(the unpruned context is $\simeq 20k$ characters towards the end).
If default tool calls fail,
(often the case for Qwen3-Coder),
we use
\textit{tool emulation}, forcing JSON outputs.
Diagnostics for GPT-4o API keys and Qwen3 server connections help resolve authentication
and network issues, respectively. 
The context truncation limit
($10k$ characters) and the head/tail size of 
standard output and standard error ($20$ lines) can also be set via code.

\subsection{Backend and frontend}
\label{subsec:backfront}

Our backend is in pure Python and does not use any
special agent libraries. The frontend is built with Streamlit\footnote{\url{https://streamlit.io}}. \cedar can run on any 
laptop, as long as the RAM allows for loading the data in main memory.
Qwen3-coder 30B is hosted on a GPU server
(4x48GB NVIDIA Ada 6000 RTX, 512 GB RAM, 64 virtual cores).

\section{Concluding remarks}
\label{sec:confut}

Through our contribution, we show that building an agentic system pays off in the long run:
the role of the human data scientist shifts from tediously scripting repetitive workflows
to more cognitively rewarding tasks like structuring requirements, scrutinizing solutions, and suggesting optimizations.

Nevertheless, our focus here was on context engineering, and agent complexity in \cedar is still rudimentary.
Natural next steps
would introduce independent agents that inspect a solution
for faithfulness with respect to original user intent,
and critique generated solution workflows for improving target metrics.

\section*{Acknowledgements}

This work has been funded by the Free State of Bavaria in the DSgenAI project (Grant No.: RMF-SG20-3410-2-18-4).
We thank members of the NLP team at Fraunhofer IIS for useful inputs at various stages of
this work.

\bibliographystyle{icml2025}
\bibliography{2026_arxiv_cedar}

\end{document}